\documentclass[conference, a4paper]{IEEEtran}
\IEEEoverridecommandlockouts
\usepackage{cite}
\usepackage{amsmath,amssymb,amsfonts}
\usepackage{url}
\usepackage{hyperref}
\usepackage{subcaption}
\usepackage{booktabs}
\usepackage{array}
\usepackage{algorithmic}
\usepackage{graphicx}
\usepackage{textcomp}
\usepackage{xcolor}
\usepackage{geometry}
\def\BibTeX{{\rm B\kern-.05em{\sc i\kern-.025em b}\kern-.08em
    T\kern-.1667em\lower.7ex\hbox{E}\kern-.125emX}}
\begin{document}

\title{Deep Active Perception for Object Detection using Navigation Proposals}

\author{\IEEEauthorblockN{Stefanos Ginargiros, Nikolaos Passalis and Anastasios Tefas}
\IEEEauthorblockA{\textit{Computational Intelligence and Deep Learning Group, AIIA Lab} \\
\textit{Department of Informatics, Aristotle University of Thessaloniki}\\
\textit{Thessaloniki, Greece }\\
\{gstefanos, passalis, tefas\}@csd.auth.gr}
}

\maketitle

\begin{abstract}
Deep Learning (DL) has brought significant advances to robotics vision tasks. However, most existing DL methods have a major shortcoming - they rely on a static inference paradigm inherent in traditional computer vision pipelines. On the other hand, recent studies have found that active perception improves the perception abilities of various models by going beyond these static paradigms. Despite the significant potential of active perception, it poses several challenges, primarily involving significant changes in training pipelines for deep learning models. To overcome these limitations, in this work, we propose a generic supervised active perception pipeline for object detection that can be trained using existing off-the-shelf object detectors, while also leveraging advances in simulation environments. To this end, the proposed method employs an additional neural network architecture that estimates better viewpoints in cases where the object detector confidence is insufficient. The proposed method was evaluated on synthetic datasets - constructed within the Webots robotics simulator -, showcasing its effectiveness in two object detection cases.

%In this paper we evaluate the effectiveness of object detection models to be able to be used in different domains when we provide a small amount of training data. First, we try the easiest and most commonly used few-shot object detection (FSOD) training scheme, that is, to use a pretrained model and then fine-tune only some of the last layers of our model. Further more we introduce our Symmetric Fine-Tuning scheme that introduce useful information faster into the trained part of our model without extra complexity. Our results indicate that good performance can be achieved in a big variety of different classes along with easier training that overcomes overfitting without big effort.
\end{abstract}


\section{Introduction}
\label{sec:intro}
Deep Learning (DL) has significantly advanced robotics vision tasks in recent years, including object detection and recognition~\cite{redmon2016you}, scene segmentation~\cite{yu2018bisenet}, face recognition~\cite{deng2020retinaface}, and more. These breakthroughs have led to impressive applications such as autonomous cars, drones, and robots that can collaborate with humans on various tasks. However, most DL methods rely on a static inference paradigm that ignores the fact that robots can interact with their environment to gather more information.
One way to address this limitation is through active perception~\cite{bajcsy1988active,bajcsy2018revisiting,passalis2022opendr}, which involves manipulating the robot or sensor to obtain a clearer view or signal. For example, adjusting a robot's pose in three-dimensional space can often improve its ability to detect objects by taking advantage of different perspectives and lighting conditions. This process is similar to how humans and animals interact with their environment, such as humans looking from different angles to process complex visual stimuli, or animals rotating their ears towards an audio source~\cite{heffner1992evolution}.
By incorporating active perception techniques into DL models, robots can improve their situational awareness and better adapt to dynamic environments. The proposed method shows potential for improving object detection capabilities in dynamic environments and could lead to advancements in the field of robotics.

Recent studies have shown that active perception can significantly improve the perception abilities of various models. One promising approach is reinforcement learning, which can be used to train deep learning models for active perception. This is demonstrated in~\cite{ammirato2017dataset}, where factors such as viewing angle, occlusions, and object scale can greatly affect recognition accuracy. Similar results have been reported in more recent works~\cite{xia2018gibson, han2019active, ramakrishnan2018sidekick}. It is worth noting that active perception approaches can often also lead to the creation of faster and lighter DL models, as they are trained to solve a simplified problem, such as face recognition from specific viewpoints~\cite{9287085}.  Despite the significant potential of active perception,  they also pose several challenges, primarily a significant change in training pipelines for deep learning models. Some approaches often require dedicated datasets to support active perception e.g., ~\cite{9287085}, which can be expensive to collect. At the same time, others rely on reinforcement learning to learn an active perception policy~\cite{ammirato2017dataset,tosidis2022active}, which is difficult to train and suffers from low data efficiency. These challenges make it difficult to implement generic active perception-enabled deep learning models that can be used for a variety of different applications.

% PRE EDIT
% To overcome these limitations, in this work, we propose a generic supervised active perception pipeline for object detection that can be trained using existing off-the-shelf object detectors. To this end, the proposed method employs an additional neural network architecture that generates movement proposals when the confidence of the object detector is insufficient. To train the additional neural network architecture, we propose constructing a confidence manifold of the object detector using a 3D simulation environment. Specifically, we directly train the navigation regressor to guide the robot towards regions with higher object detection confidence, allowing it to actively perceive and explore the environment. At the same time, the risk of distribution shifts~\cite{ho2021retinagan} due to the use of a simulator can be mitigated to some extend, since the original object detector is not required to be trained in simulation, while the use of supervised learning instead of reinforcement learning makes this process much easier to apply in practice. The proposed method is evaluated using synthetic datasets constructed in the Webots robotics simulator, demonstrating its potential in two object detection cases. An open source implementation of
% the proposed method is available at: \url{URL-REMOVED-FOR-PEER-REVIEW}.

To overcome these limitations, our proposed method employs an additional neural network architecture that generates movement proposals when the confidence of the object detector is insufficient. To train this architecture, we propose constructing a confidence manifold of the object detector using a 3D simulation environment. Specifically, we train a navigation regressor to guide the robot towards regions with higher object detection confidence, enabling the system to actively perceive and explore the environment.
By internally representing the confidence manifold of a variety of object classes -through training- in the Navigation Network, our approach can efficiently learn to explore and perceive the environment in a targeted manner. At the same time, the risk of distribution shifts~\cite{ho2021retinagan} due to the use of a simulator can be mitigated to some extend, since the original object detector is not required to be trained in simulation, while the use of supervised learning instead of reinforcement learning makes this process much easier to apply in practice.

The rest of the paper is structured as follows. The proposed method is presented in Section~\ref{sec:proposed}. Then, the experimental evaluation is provided in Section~\ref{sec:eval}. Finally, conclusions are included in Section~\ref{sec:conclusions}.

\section{Proposed Method}
\label{sec:proposed}

In this section, we present a high-level summary of the active perception approach pipeline, while also delving into the specific steps involved in training the navigation proposal network that forms a key part of this approach.

% PRE EDIT
% \begin{figure}
% \centering
% \includegraphics[width=0.99\linewidth]{figs/complete.png}
% \caption{Active perception pipeline for object detection. First, we employ a regular object detector. If we are uncertain regarding the object that has been detected, then we employ a navigation network to acquire a navigation proposal. During the execution of this proposal, we also employ object detection at predefined points as shown in Fig.~\ref{fig:auth-outer_movement}. At the end of the execution of the navigation proposal, we have obtained a better view of the object at hand.}
% \label{fig:auth-netw}
% \end{figure}

\begin{figure}
\centering
\includegraphics[width=0.99\linewidth]{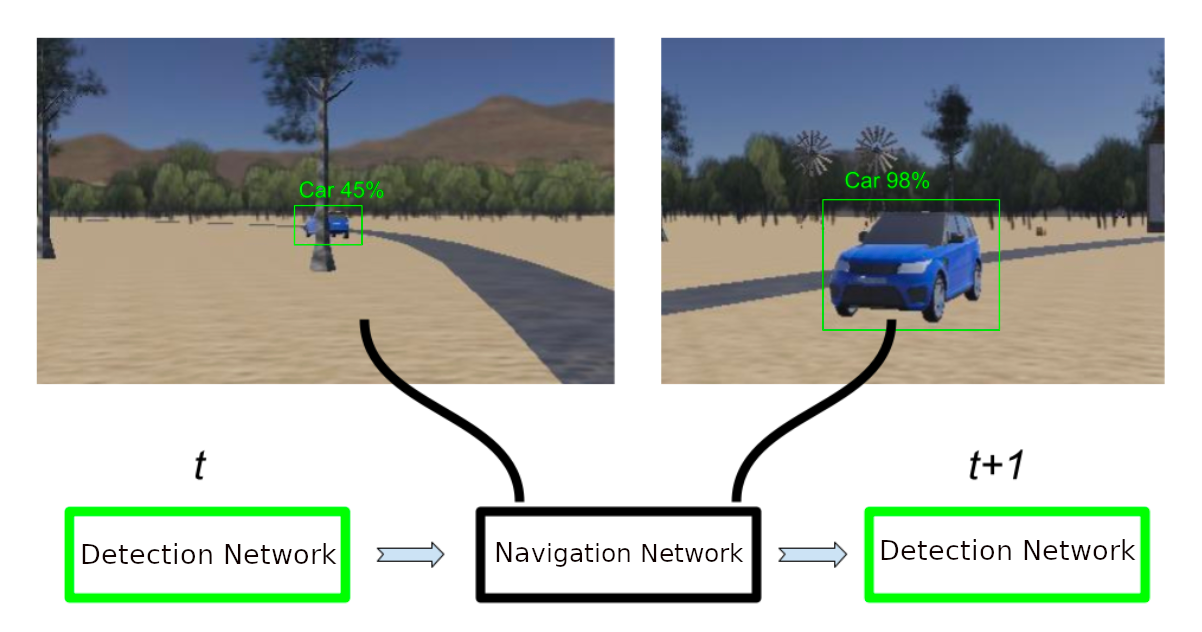}
\caption{Illustrates the execution of a navigation proposal in the proposed active perception pipeline. If the object detector's confidence score falls below a threshold, indicating uncertainty regarding the detected object, the navigation network generates a navigation proposal that directs the robot to move towards regions with potentially higher object detection confidence. During the execution of the navigation proposal, object detection is also performed at predefined points along the path as shown in Fig.~\ref{fig:auth-inner_movement}.}
\label{fig:auth-netw}
\end{figure}

\subsection{Active Perception Pipeline}
The proposed pipeline is outlined in Fig.~\ref{fig:auth-netw}. First, (i)  object detection is executed and the confidence of the object of interest is evaluated. If the detection is below a desired threshold, e.g., 80\%, then (ii) the navigation module triggers and proposes a position/rotation in the 3D space. After executing this navigation proposal, an updated view is obtained that can lead to increased object detection accuracy. In the remainder of this subsection we provide details for each of the employed steps.
\\ \\
\begin{figure}[t]
\centering
\includegraphics[width=0.95\linewidth]{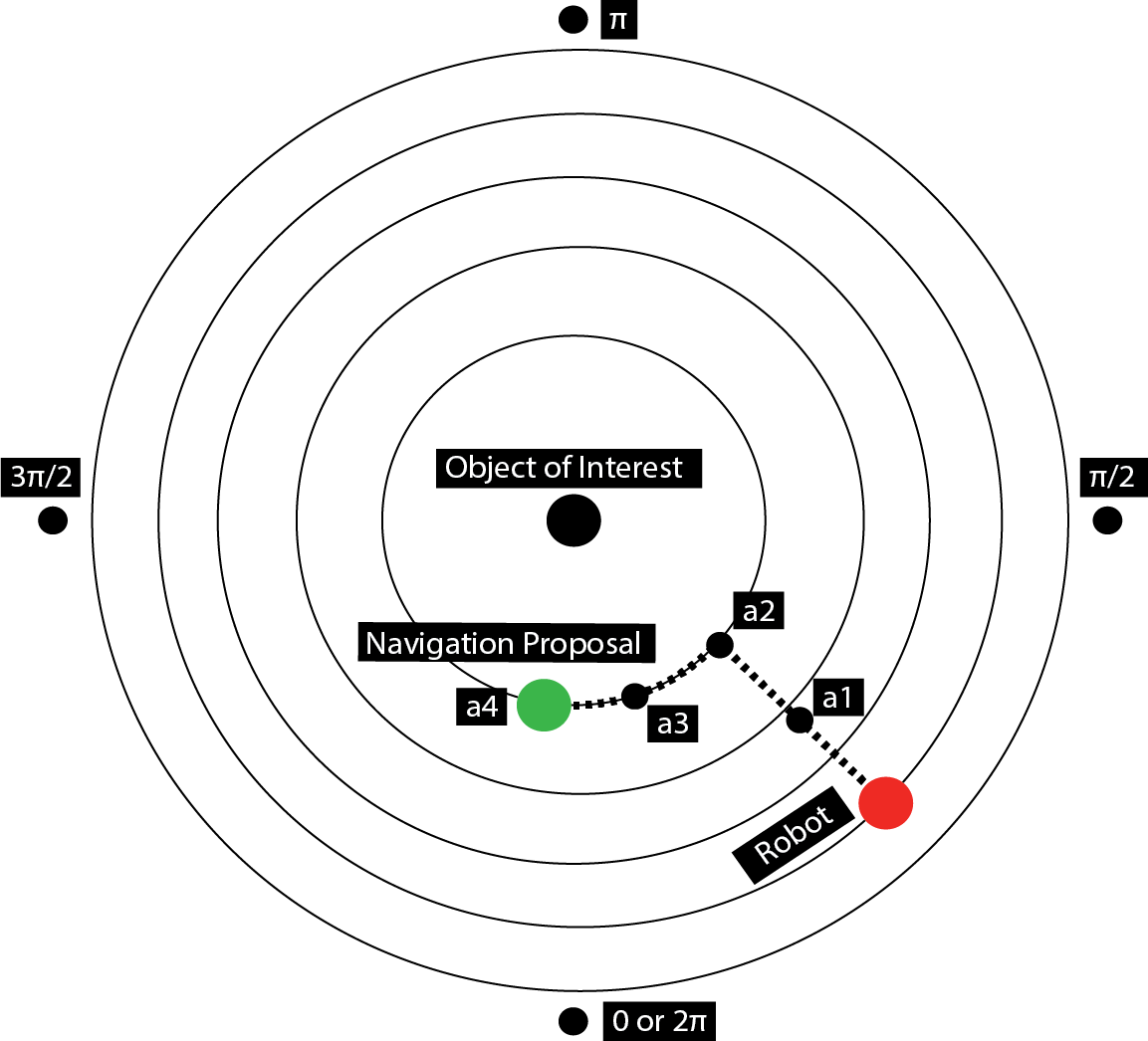}
\caption{Movement example of a robot that performs active perception in order to increase object detection confidence, while moving towards the object (inwards movement). Note the intermediate points ($a_1$ to $a_4$) where object detection is re-employed. Red point: Initial Position, Green point: Navigation Proposal. }
\label{fig:auth-inner_movement}
\end{figure}
\begin{figure}[t]
\centering
\includegraphics[width=0.95\linewidth]{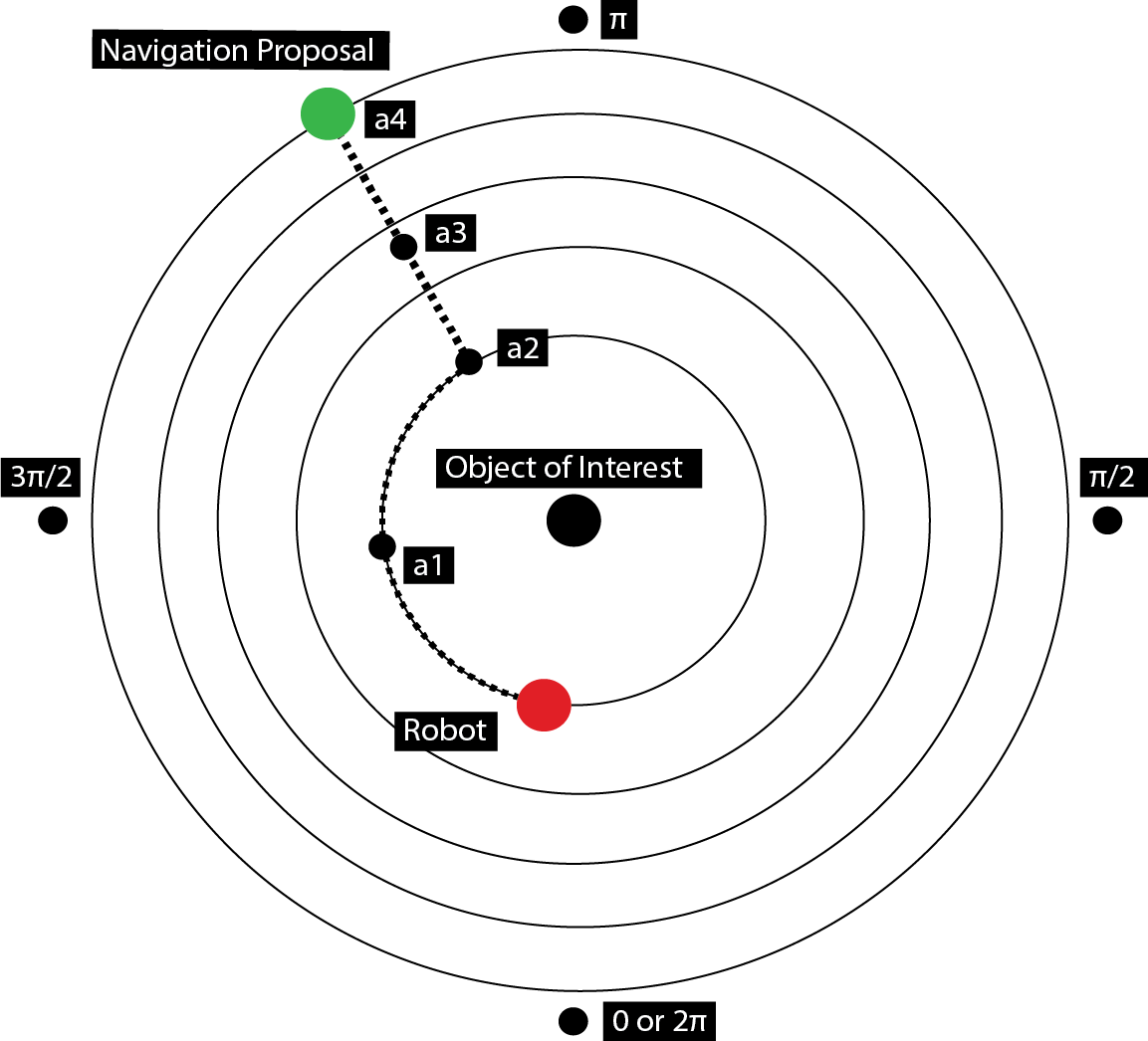}
\caption{Movement example of a robot that performs active perception in order to increase object detection confidence, while moving away from the object (outwards movement). Note the intermediate points ($a_1$ to $a_4$) where object detection is re-employed. Red point: Initial Position, Green point: Navigation Proposal. }
\label{fig:auth-outer_movement}
\end{figure}
\underline{\textbf{Phase 1: Initial Object Detection}}
\\
In this step, a regular deep learning-based object detector is employed. This detector can be described as a function $f_{detect}(\mathbf{x}_i$), where $\mathbf{x}_i$ is the observation at the $i$-th time step. The observation in this particular case is a 2D projection (a camera image) of the 3D world in which the robot resides. The detector produces a confidence score $p_i$ which describes the certainty of the appearance of the object of interest in the observation $\mathbf{x}_i$, as well as its location and size $(y_h, y_v, w, h)$ in the 2D input space (image).
If the confidence score ${p}_i$ of the object of interest is below a threshold (i.e., if the detection is inadequate), then we proceed with \emph{Phase 2} and \emph{Phase 3} in order to improve the results. Note that employing correctly calibrated object detectors for uncertainty estimation is critical. Therefore, uncertainty quantification~\cite{abdar2021review} and confidence calibration~\cite{mehrtash2020confidence} approaches can be employed for this purpose.
\\ \\
\underline{\textbf{Phase 2: Navigation Proposal}}
\\
The subsequent step entails identifying the optimal robot movement that can maximize the object detection confidence score. To this end, in this work we employ a dedicated \textit{navigation network}. A convolutional neural network designed for regression tasks that can estimate the optimal navigation plan by maximizing the confidence score. In other words, we try to maximize the confidence ${p}_{i+1}$ of the object detector at the {next} observation $\mathbf{x}_{i+1}$ at the next time step $i+1$. The navigation module is a function that takes the observation $\mathbf{x}i$ as input and outputs the optimal translation and rotation vectors $\mathbf{z}_{i+1}$ and $\mathbf{r}_{i+1}$ for the robot's movement, respectively. These vectors direct the robot to the \emph{next} optimal point in the 3D space by utilizing distilled knowledge from the confidence manifolds that has been trained on. The final step of this phase involves applying these transformations to the robot. The optimal rotation vector $\mathbf{r}_{i+1}$ generated by the navigation module directs the robot to rotate around the object of interest, while the translation vector $\mathbf{z}_{i+1}$ determines the direction and distance of the robot's frontal or backward movement, 
\\ \\
\underline{\textbf{Phase 3: Object Detection in Trajectory}}
\\
Given the estimated translation vector $\mathbf{z}_{i+1}$ and rotation vector $\mathbf{r}_{i+1}$ we can formulate a trajectory from the starting point at time step $i$ to the ending (optimal) point at time step $i+1$. As the robot moves along the computed trajectory, object detection can be performed multiple times to evaluate the object detection confidence score until it reaches the final destination, thereby further improving the perception accuracy of the system. An example of such a robot movement is depicted in Fig.~\ref{fig:auth-inner_movement}. Similarly, the proposed method can be applied to perform movements where the robot moves away from the object of interest, as shown in Fig.~\ref{fig:auth-outer_movement}.

\subsection{Navigation Proposal Network Training}
In order to train the navigation proposal network $f_{nav}(\mathbf{x}_i)$, acquiring ground truth annotations is necessary. Without loss of generality, the proposed approach can be first applied to the one-dimensional case, where the navigation proposals involve navigation along a single axis, such as the angle between the robot and the object as the robot moves on a circular trajectory around the object. In this case, the confidence of the object detector can be exhaustively evaluated for each discrete point on the circle after quantization. Then, for each angle $\theta$ there is a corresponding observation $\mathbf{x}_\theta$, as well as a confidence $p_\theta$, acquired from object detection  $f_{detect}(\cdot)$. For each of these observations, a navigation proposal can be generated, i.e., a target angle $t_\theta$, which quantifies robots rotation with the object of interest as pivot point. Essentially this process is identifying the nearest point on the circle where the object detection confidence score exceeds a predefined threshold $p_{thres}$. The Navigation Proposal Network can be effectively trained by minimizing the regression error between the ground truth navigation proposals and the network's estimated proposals. This can be achieved by formulating the navigation loss $\mathcal{L}_{nav}$ as follows:
\begin{equation}
    \mathcal{L}_{nav} = (f_{nav}(\mathbf{x}_\theta) - t_{\theta})^2.
    \label{eq:loss}
\end{equation}
To minimize $\mathcal{L}_{nav}$, we employ the widely used back-propagation technique. This involves computing the gradients of the objective function with respect to the network's parameters, and updating the parameters using an optimization algorithm.
Note that the proposed method can be trivially extended to handle 3D control, as demonstrated in the experimental evaluation, by employing the $l_2$ distance between the target and regressed rotation and translation vectors, and by extending the ground truth generation process to additional axes. This process is also further explained in Section~\ref{sec:eval}, including both a 2D example (Fig.~\ref{auth:2D-Labels}), as well as 3D evaluation (Fig.~\ref{auth:3D-Eval} and~\ref{auth:3D-Eval-person}).

\section{Experimental Evaluation}
\label{sec:eval}

The proposed method was evaluated using the Webots simulation environment, using a DJI Mavic drone as a robot. A core controller for the drone has been developed, which offers essential functionalities related to movement control and sensor data acquisition. More specifically, the drone can be moved in two distinct ways, either by utilizing the Supervisor class in Webots \cite{b16}, which allows instant translation/ rotation anywhere in the three dimensional space bypassing the physics of the simulation, or by fully activating the drone motors and the atmosphere/gravity in the environment to test for real-world conditions.
Furthermore, a camera system has been implemented on the drone to enable visual data acquisition. The camera captures data in 30 frames per second. 
The developed simulation environment has been used to create two distinct datasets for training the employed deep learning models: a classic object detection dataset and an active perception dataset that supports the corresponding task. Simulated 3D models of a Tesla Model S and a Toyota Corolla, as well as realistic 3D human models and dolls, were used to conduct the experiments inside the Webots simulation environment. All of the synthetic data were created using aforementioned method. These datasets were then utilized to fine-tune pre-trained object detectors within this environment, as well as train the custom navigation network from scratch. For the regular object detection dataset, images of cars and persons were extracted in different world settings and angles. The object detector was fine-tuned using a total of 5,000 images captured at 65 different radius values and 76 angles around the objects of interest. The additional training/finetuning step was undertaken to enhance the performance of the object detector for our specific use case, instead of relying on a generic trained object detector. Although optional, this step was deemed necessary to prevent potential distribution shifts during the evaluation of the object detector. Each image in the dataset is labeled with a pair of rotation and translation vectors that direct the drone to a viewpoint in the 3D space that provides a more informative observation for the subsequent image, as described in Section~\ref{sec:proposed}. The max distance from the objects was 60 meters.

The proposed architecture consists of two distinct neural networks, namely the object detection network and the navigation proposal network. The object detection network continuously scans the environment for objects, specifically cars or persons in this experiment, and performs adequately when the detection confidence exceeds a predefined threshold. When the object detector fails to identify the objects with sufficient accuracy, the navigation proposal neural network takes control to suggest a movement in the 3D space that enhances object detection performance. Notably, both neural networks use the same input data, which comprises the video stream frames captured by the drone's camera. In this work, we use an SSD-based architecture for object detection \cite{b13} with the addition of a VGG feature extraction layer~\cite{sengupta2019going}. Furthermore a set of auxiliary convolutional layers (from conv6 onwards) were added, allowing features to be extracted at multiple scales and the size of the input to each subsequent layer to be progressively reduced. The navigation module employed a ResNet-18 convolutional architecture, as proposed in the work by He et al. \cite{b18}. The input of the network is adjusted to fit the dimensions of the camera input stream, i.e., (420 $\times$ 240 $\times$ 3), whereas the output dimensions were configured to regress the rotation and translation vectors. The sigmoid activation function was used for the output layer of the network. Note that the training targets were appropriately normalized to support this architecture, i.e., they were normalized to (0, 1). The navigation proposal network was trained using the Adam optimizer~\cite{kingma2014adam}, using a learning rate of 0.001 and its default hyperparameters.

\begin{figure}[t]
\centering
\includegraphics[width=0.99\linewidth]{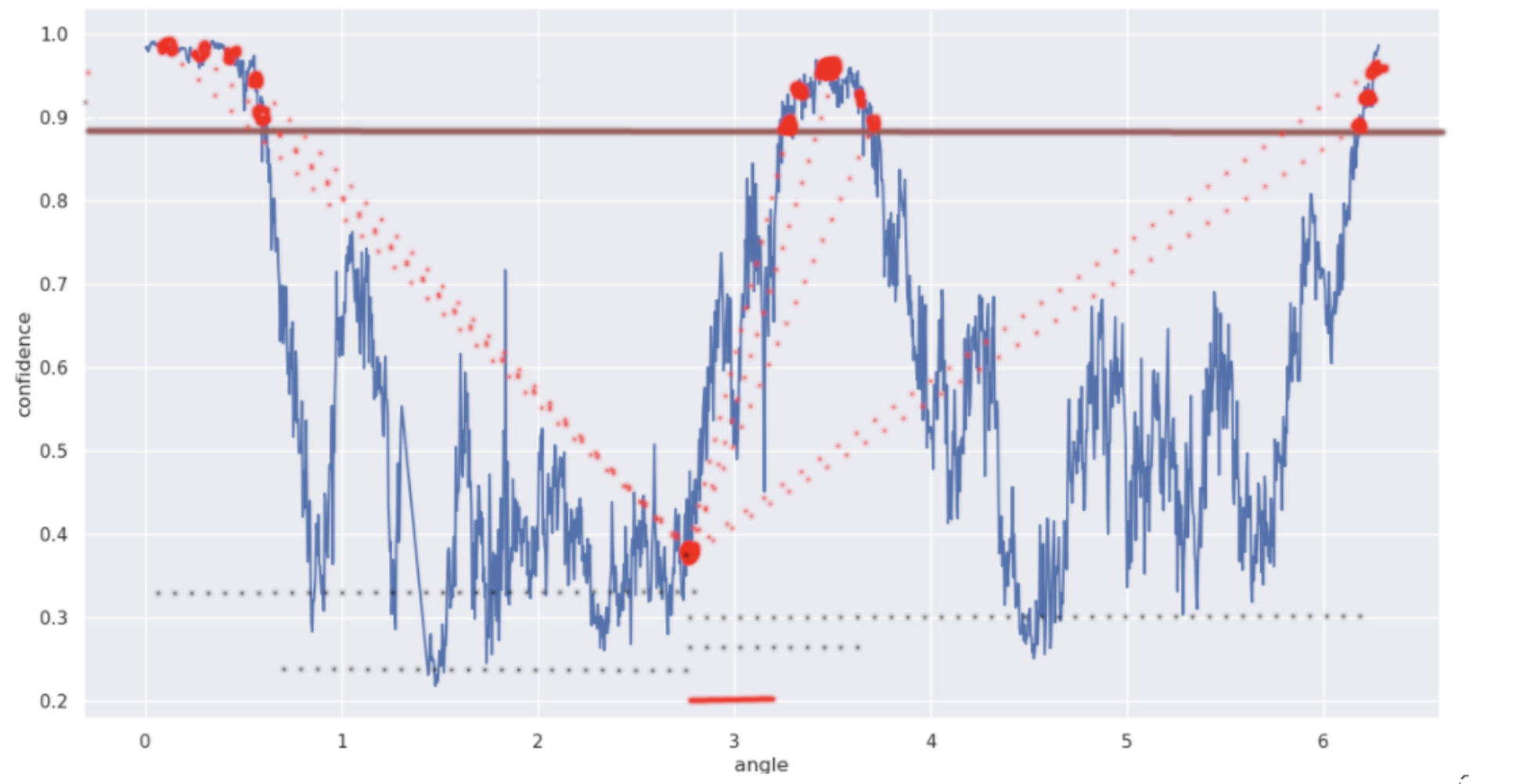}
\caption{Mapping different position to optimal navigation plan according to object detection confidence}
\label{auth:2D-Labels}
\end{figure}

\begin{figure}[t]
\centering
\includegraphics[width=0.99\linewidth]{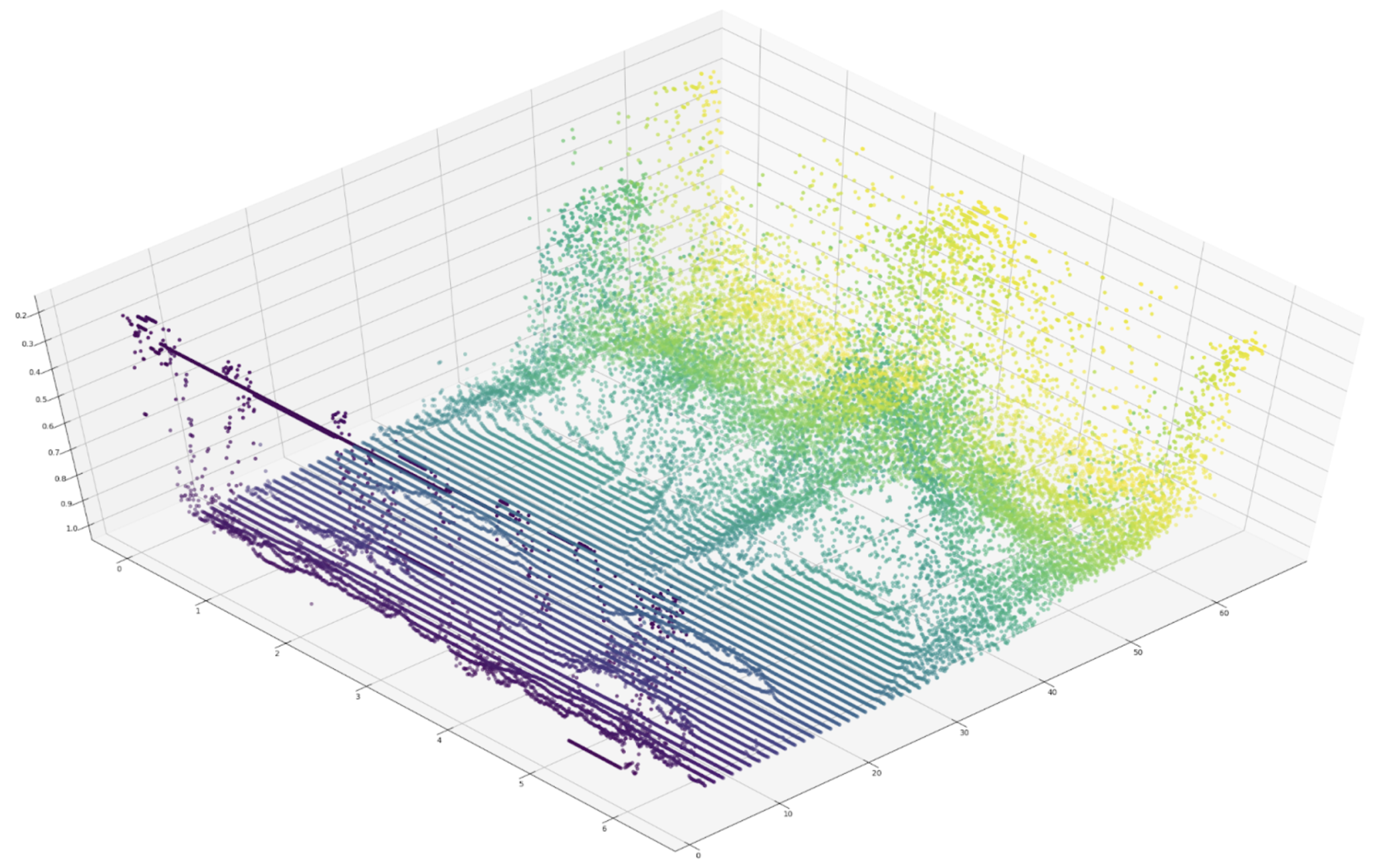}
\caption{Car detection confidence at different angles and distances}
\label{auth:3D-Eval}
\end{figure}

\begin{figure}[t]
\centering
\includegraphics[width=0.99\linewidth]{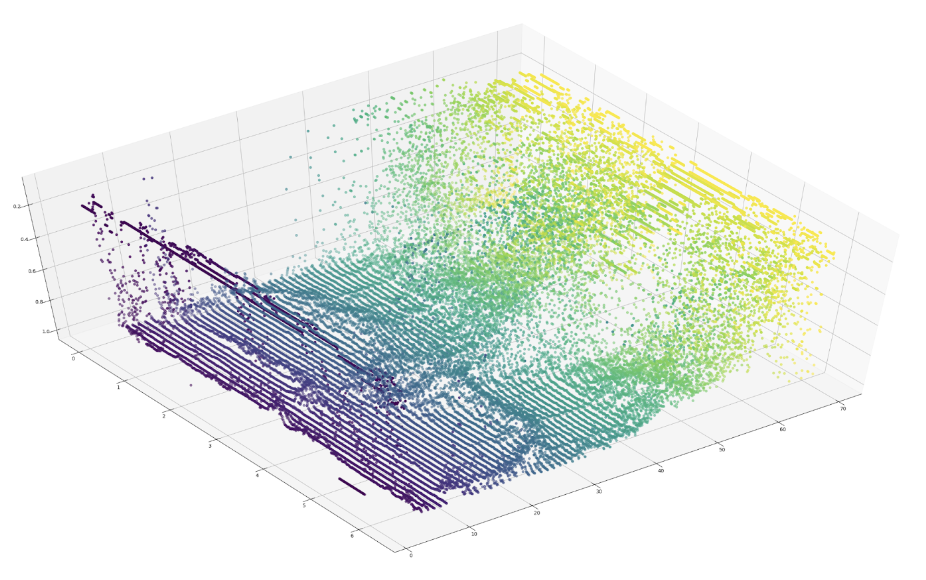}
\caption{Person detection confidence at different angles and distances}
\label{auth:3D-Eval-person}
\end{figure}

\begin{figure*}[th!]
  \centering
  \begin{minipage}{0.32\textwidth}
    \includegraphics[width=\textwidth]{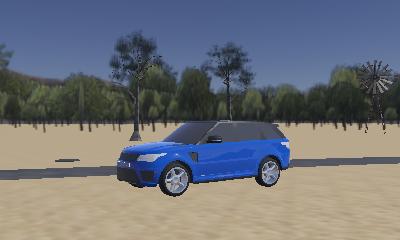}
  \end{minipage}
  \begin{minipage}{0.32\textwidth}
    \includegraphics[width=\textwidth]{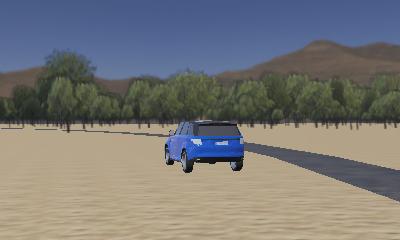}
  \end{minipage}
  \begin{minipage}{0.32\textwidth}
    \includegraphics[width=\textwidth]{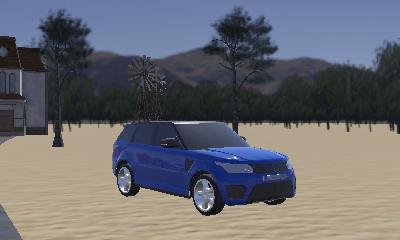}
  \end{minipage}
  \caption{Random initialization during evaluation}
  \label{auth:init-eval}
\end{figure*}

The complete active perception pipeline comprises both the object detection and navigation proposal networks. Initially, the object detection network runs to detect the presence of objects and computes the corresponding confidence scores. These confidence scores are subsequently compared against a predefined threshold to assess the accuracy of the detection. In the context of the navigation proposal network dataset, the thresholds used were set to 0.7, 0.8, 0.9, 0.95, and 0.98. We achieved the best results using the $p_{thres}=0.9$ threshold for the dataset creation. Thus, for our use case when the detector detects an object with confidence below $p_{thres}=0.9$ accordingly, the navigation network is enabled.  

By leveraging the object detection evaluations detailed earlier, a critical set of points in the 3D environment was quantized and mapped to another set of points that optimized the viewpoint detection confidence. Subsequently, the navigation labels were generated based on these points. To better illustrate the ground truth generation process, as seen in Fig \ref{auth:2D-Labels} these points are mapped to the closest possible points near them over a viewpoint confidence threshold. In this specific example, the closest point to the current position was ultimately mapped as the optimal viewpoint. The red bar below Fig.~\ref{auth:2D-Labels} represents the closest point distance from the best viewpoint. The label for this example would be the distance from the best viewpoint signed positively (meaning rotation around the object to the right), which would be +0.40 rads. In the case of 3D navigation, a similar approach is applied by including an additional dimension for translation movement, while preserving the methodology used for 2D navigation.

Figure~\ref{auth:3D-Eval} depicts the confidence of an object detector, used for car detection, at various angles and distances, providing an illustration of the process of dataset creation for object detection in the 3D case.  Additionally, Figure~\ref{auth:3D-Eval-person} displays the evaluation results for person detection, highlighting the distinct probability manifolds for each object class, which are determined by their intrinsic characteristics (symmetries, etc.).

\begin{table}[]
\caption{Experimental Evaluation (Success rate 
 and improvement rate). The improvement rate concerns only the successful runs (otherwise the improvement is 0\%).}
    \label{tab:my_label}
    \vspace{1em}
    \centering
    \begin{tabular}{c|cc}
    \toprule\\
    \textbf{Methods} & \textbf{Success Rate} & \textbf{Improv. Rate} \\
    \midrule
     \textbf{Static Perception }&  0\% & 0.0\% \\
     \textbf{Random Navigation}           & 22\% & 60.6\%\\
     \textbf{Clas.-based Nagivation}   & 51\% & 66.8\%\\
     \textbf{Proposed}         & \textbf{67\%} & \textbf{70.4\%}\\     
     \bottomrule
    \end{tabular}
    
\end{table}

To evaluate the proposed active perception pipeline, 100 experiments were conducted for each evaluation case and the total improvement of the object detector was averaged. Testing was performed in an unknown environment for both models. In each experiment, the robot's position-rotation was randomized, allowing it to spawn in different coordinates of the 3D map, as shown in Fig.~\ref{auth:init-eval}. The experimental results are reported in Table~\ref{tab:my_label}. The success rate and improvement rate were both measured in this study. The success rate refers to the number of evaluated cases where active perception improved the obtained results. The improvement rate quantifies how much the prediction confidence was enhanced for the cases where active perception was successful. The proposed method manages to increase the object detection confidence in 67\% of the evaluation cases, i.e., in 67\% of the evaluated cases the final object detection confidence was higher than the initial object detection confidence.  A random navigation policy led to a significantly lower improvement rate of 22\%.  It is worth noting that a random exploration policy still leads to improvements due to the intermediate snapshot evaluations that are performed. The intermediate points that do not lead to improvements are simply discarded, so the random navigation policy can lead to improvements in the cases where some of these points happen to increase the detection confidence. Finally the classification navigation policy was also evaluated, which involves selecting the direction of movement through classification rather than regression of the actual movement. i.e., by extending~\cite{9287085} to match the used setup. In this case, the active perception policy increased the confidence in 51\% of the evaluated cases, further highlighting the improvements obtained using the proposed method.

\section{Conclusions}
\label{sec:conclusions}

% In this work, we presented a generic supervised active perception pipeline for object detection that can be trained using existing off-the-shelf object detectors. The proposed method uses an additional neural network architecture that generates movement proposals when the confidence of the object detector is low. The proposed method was evaluated using synthetic datasets constructed in the Webots robotics simulator, demonstrating its potential in two object detection cases, i.e., car detection and person detection.  The conducted experiments demonstrated that it is possible to train DL models to regress a robot plan that can increase object detection confidence, providing an easy and practical way to equip robots with active perception capabilities.  The proposed method can be readily extended to cover other cases for which pre-trained detectors and simulation models exist. This promising result can be further explored by incorporating the regression network into the object detection model, providing active perception  at virtually no cost, as well as exploiting co-integrated simulation and training in order to reduce the sim-to-real gap and ensure that models will perform as expected in real conditions.

In this work, we presented a generic supervised active perception pipeline for object detection that can be trained using existing off-the-shelf object detectors. The proposed method was evaluated using synthetic datasets constructed in the Webots robotics simulator, demonstrating its potential in two object detection cases, i.e., car detection and person detection. This approach can be readily extended to various object detection scenarios where trained detectors and simulation models are available. It offers a practical and straightforward means to endow robots with active perception capabilities. The promising results suggest further exploration of incorporating the regression network into the object detection model, providing active perception at virtually cost, and leveraging co-integrated simulation and training to reduce the sim-to-real gap.

\section*{Acknowledgment}
This work was supported by the European Union’s Horizon 2020 Research and Innovation Program (OpenDR) under Grant 871449. This publication reflects the authors’ views only. The European Commission is not responsible for any use that may be made of the information it contains.

\bibliographystyle{ieeetr}
\bibliography{bib}

\end{document}